%% file: iclr2026_conference.tex
\documentclass{article} 
\usepackage[T1]{fontenc}
\usepackage{iclr2026_conference,times}

\input{math_commands.tex}

\usepackage{hyperref}
\usepackage{url}
\usepackage{graphicx}
\usepackage{wrapfig}
\usepackage{tcolorbox}
\usepackage{listings}

\title{Programming by Backprop: An Instruction is Worth 100 Examples When Finetuning LLMs}

\author{%
  Jonathan Cook\thanks{Equal contribution.} \\
  FLAIR, University of Oxford\\
  \texttt{jonathan.cook2@hertford.ox.ac.uk} \\
  \And
  Silvia Sapora$^*$ \\
  FLAIR, University of Oxford \\
  \texttt{silvia.sapora@stats.ox.ac.uk} \\
  \AND
  Arash Ahmadian\thanks{Now at Google DeepMind.} \\
  Cohere \& Cohere Labs \\
  \And
  Akbir Khan \\
  Anthropic \\
  \And
  Tim Rocktäschel \\
  UCL AI Centre \\
  \And
  Jakob Foerster \\
  FLAIR, University of Oxford \\
  \And
  Laura Ruis \\
  MIT \\
  \texttt{lruis@mit.edu} \\
}

\iclrfinalcopy 
\begin{document}

\maketitle

\begin{abstract}
Large language models (LLMs) are typically trained to acquire behaviours from demonstrations or experience, yet much of their training data is declarative: instructions, rules, and descriptions that specify behaviours without showing how to execute them. We introduce \textbf{Programming by Backprop (PBB)}: a training regime that enables LLMs to acquire \emph{procedural} knowledge (i.e., reusable behaviours) from \emph{declarative} instructions encountered during training. With PBB, instructions in training data provide an opportunity to `program' specific behaviours into model weights. The core principle underpinning PBB is the separation of learning how instructions map to behaviour from internalising new instructions. We devise two distinct PBB curricula that leverage this principle. Through controlled experiments across two domains (algorithmic execution from Python source code and text generation from context-free grammars), we demonstrate the benefit of these curricula over training on a homogeneous data mixture. Crucially, PBB is highly sample efficient, with \emph{a single instruction substituting for up to 100 execution examples}. Though execution of instructions in training data remains less reliable than when instructions are given in-context, our results demonstrate that procedural knowledge can be noisily `programmed' into LLMs through PBB, with important implications for data curation and safety.
\end{abstract}

\section{Introduction}
\label{intro}

\begin{figure*}[t]
\begin{center}
\includegraphics[width=0.95\columnwidth,trim={22cm 47.0cm 45cm 16cm}, clip]{"figures/pbb_new1_main.pdf"}
\end{center}
\caption{Illustration of \emph{Programming by Backprop} (PBB) --- the learning of behaviours from instructions. We define train instructions as those for which examples of behaviour are also available. Evaluation instructions (yellow in the figure) correspond to the behaviours being tested, which are never demonstrated in training data. PBB comprises two training curricula: \textbf{Proactive PBB}, where models learn a general correspondence between instructions and behaviours before being exposed to evaluation instructions, and \textbf{Retroactive PBB}, where initial exposure to all instructions is followed by learning how to map those to behaviour using the train set.}
\label{fig:opener}
\end{figure*}

Large language models (LLMs) are typically trained to acquire behaviours from demonstrations, via pretraining and supervised finetuning (SFT), or from experience, via reinforcement learning (RL). Yet much of the data that models are exposed to, particularly during pretraining, does not consist of demonstrations, but of abstract instructions, rules, algorithms, and descriptions that specify procedures without showing how they execute on concrete inputs. For humans, such symbolic knowledge plays a central role in learning general skills, allowing abstract instruction to complement demonstration and practice, substantially improving learning efficiency \citep{Dienes1999Theory}. Whether LLMs can similarly acquire procedural knowledge from declarative training data remains an open question. 

Users of instruction-tuned LLMs commonly give instructions in-context through prompting, meaning that the model's generation behaviour is explicitly conditioned on that instruction. By contrast, internalising behaviour from instructions in \emph{training} data would amortise the per-instance computation associated with conditioning on that description into the model's parameters, enabling a model to reuse the behaviour in a range of contexts and compose it with other learned behaviours. Understanding how training on instructions can influence the downstream behaviour of an LLM is crucial both for safety and for developing models with skills beyond those demonstrated extensively in training data. 

\begin{wrapfigure}{r}{0.49\textwidth}
\vspace{-2em}
\centering
\includegraphics[width=0.45\textwidth]{"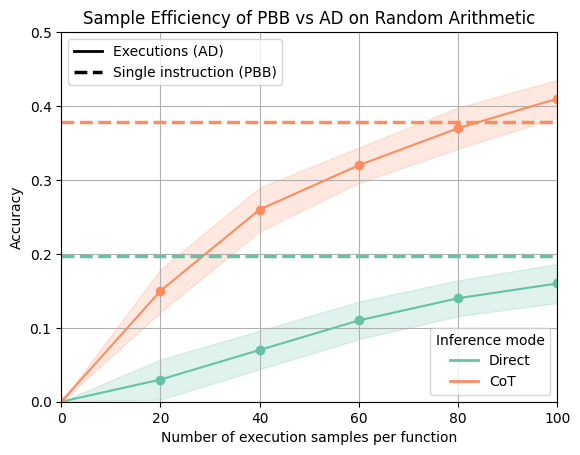"}
\vspace{-1em}
\caption{PBB enables \texttt{Qwen3-14B} to learn an algorithm from a single instruction (piece of code) to similar efficacy as using Algorithm Distillation (AD) --- learning an algorithm from input-specific examples --- on 100 examples.}
\label{fig:random_efficiency}
\vspace{-1em}
\end{wrapfigure}

We empirically demonstrate that the local objective of maximising the likelihood of an instruction does not, in isolation, incentivise learning the corresponding behaviour; rather, this acquisition relies on a form of \emph{latent learning} \citep{Lampinen2025Latent} driven by the structure of the full training data distribution. Prior work provides suggestive evidence of latent learning, showing that LLMs can emulate simple character descriptions encountered during finetuning \citep{berglund2023oocr}, extract general strategies from training examples \citep{betley2025tell}, and exhibit downstream skills correlated with the presence of input-general definitions in pretraining data \citep{ruis2025procedural}. However, existing work does not methodologically isolate whether an instruction encountered during training is sufficient to induce execution of an associated multi-step, input-conditional behaviour. Nor does prior work identify approaches that improve an LLM's ability to internalise behaviour from instructions. 

We introduce \textbf{Programming by Backprop (PBB)} as a training approach for inducing context-dependent, executable behaviours from abstract instructions. In PBB, behaviours on which the model will be evaluated are specified declaratively in training data, rather than demonstrated, and training is designed to compile these specifications into the model’s parameters. In this sense, `programming' refers to the explicit inclusion of procedural instructions in training data, together with training strategies that induce the corresponding behaviour at inference time.

PBB is instantiated through structured, multi-stage curricula that combine instruction modelling with execution supervision over a designated train set of behaviours. We design two such curricula. In the first, models are trained to learn a general correspondence between instructions and behaviours using train behaviours, and then apply this correspondence to new instructions corresponding to test behaviours introduced in a subsequent stage. This regime is a form of implicit meta-learning as introduced by \cite{Krasheninnikov24}, and closely mirrors their two-stage training pipeline. In the second, models are first trained on the full set of instructions, acquiring declarative knowledge without execution competence; execution supervision on train behaviours is then used to convert this latent declarative knowledge into procedural knowledge in a way that generalises to test behaviours. We refer to these curricula as proactive and retroactive PBB, respectively (see Figure \ref{fig:opener} for an overview).

We evaluate PBB across two domains --- algorithmic execution from Python source code and text generation from context-free grammars --- and find that it enables highly sample-efficient learning. Under PBB, a single instruction can substitute for many demonstrations while achieving comparable performance (Figure \ref{fig:random_efficiency}). PBB can also mitigate biases arising from imbalanced demonstration data (Section \ref{subsec:results:sampleefficiency}). In addition, models trained with PBB exhibit a limited ability to compose independently trained algorithmic instructions at inference time, and we find that algorithmic execution are learned more effectively when instructions are expressed in code rather than natural language (Section \ref{sec:algorithmic_results}). Together, these results characterise how PBB can be used to program reusable behaviours into language models via training on instructions.

Finally, while finetuning with PBB induces executable behaviour, execution remains less reliable than when instructions are provided explicitly in-context. Nonetheless, our experiments show that models trained with PBB can noisily execute instructions encountered during training within the forward pass, without relying on chain-of-thought reasoning. We further observe positive scaling trends with respect to both model size and data, and note that the PBB curricula complement prior work highlighting the role of curriculum design in continual pretraining \citep{Parmar2024Reuse, Cagatay2025Investigating, Chen2025Towards, Ou2025How}. These findings suggest that PBB is a promising training methodology for further scaling.

\section{Related Work}
\label{related_work}

\paragraph{Learning to execute.}
A large body of work studies how neural models can learn to execute programs when given explicit execution supervision. Early work shows that recurrent networks can map program text to outputs given paired input–output examples \citep{zaremba2015learningexecute}. Subsequent approaches extend this direction to richer settings, such as inferring symbolic ``shape programs'' from perceptual data and executing them for downstream tasks \citep{Tian2019Learning}, or training transformer-based execution engines on demonstrations of algorithmic subroutines \citep{Yan2020Neural}. Other work introduces architectural inductive biases, such as neural interpreters, to achieve modular and systematic execution \citep{Rahaman2021Dynamic}. While conceptually related in their focus on executable behaviour, these approaches assume that procedures are either inferred from data or trained with execution-level supervision for each procedure. In contrast, PBB involves learning executable behaviours from symbolic descriptions.

\paragraph{Code training and systematic reasoning.}
Several recent works investigate how exposure to code and other symbolic structures during pretraining affects downstream model behaviour. Training on source code has been shown to improve reasoning and problem-solving in natural language tasks \citep{aryabumi2024code, Petty24}, suggesting that the regularity and compositional structure of code can scaffold general reasoning abilities. Related work shows that training on synthetic procedural traces, such as edit sequences, improves code synthesis by encouraging models to represent intermediate transformations \citep{piterbarg2025training}. Other approaches explicitly couple language models with external interpreters, grounding reasoning in code execution to improve reliability \citep{Li24chain}. This literature demonstrates that symbolic supervision can support systematic reasoning. PBB differs in that symbolic procedures are not used as external tools or reasoning aids at inference time, but instead serve as training-time programs whose execution semantics are internalised into the model parameters. Recent work has begun to investigate the kinds of pretraining data that influence reasoning capabilities. \citet{ruis2025procedural} show that exposure to input-general procedures, such as code functions, in pretraining data strongly influences models’ ability to solve related input-specific reasoning problems. This suggests that declarative data can shape downstream execution behaviour, motivating controlled studies of this process. 

\paragraph{Limits of procedural generalisation.}
Prior work on arithmetic reasoning shows that LLMs often rely on surface heuristics rather than executing explicit algorithms \citep{nikankin2025arithmetic}, though structured prompting can elicit more systematic behaviour \citep{chen2023program, Chen24}. Mechanistic analyses further suggest that the autoregressive training objective shapes which procedural abstractions can be internalised, constraining both successes and failure modes \citep{mccoy2024embers, Wang24}. Relatedly, \citet{Allen23} show that LLMs exhibit a gap between `knowing' and `doing' in tasks requiring classification or comparison, offering a potential explanation for observed failures of applying knowledge about a task to solving the task itself \citep{paglieri24}. \citet{lampinen2025generalization} further demonstrate that generalisation differs substantially between knowledge provided in-context and knowledge acquired during training. Our results contribute to this discussion by showing that data curricula can enable models to convert declarative knowledge into procedural knowledge that generalises across contexts at inference time.

\paragraph{Out-of-context reasoning.}
Finally, PBB connects to a growing body of work on how LLMs generalise from their training data in sophisticated ways. Prior studies show that models can perform out-of-context reasoning (OOCR) \citep{berglund2023oocr, betley2025tell}, where inference time behaviour is informed by training data that indirectly or abstractly describes relevant context. Models have also been shown to exhibit implicit meta-learning \citep{Krasheninnikov24}, perform latent multi-hop reasoning \citep{Yang24}, and infer latent structure about their training data \citep{Treutlein24}. PBB can be framed as a specific form of out-of-context reasoning in which models learn representations of symbolic abstractions that permit execution at inference time. 

\section{Programming by Backprop}

We formalise Programming by Backprop (PBB) as a learning regime in which the behaviour defined by an instruction is acquired by training on that instruction and encoded into a language model’s parameters for execution.

\subsection{A Programming Perspective on Training Data}

Let $\mathcal{S}$ be a space of symbolic instructions, such as Python programs or formal grammar rules. We associate each instruction $s \in \mathcal{S}$ with its denotation $[\![s]\!]$, which captures the behaviour that the instruction defines.

In this work, we consider two domains:

\begin{enumerate}
    \item \textbf{Algorithmic Execution:} For a program $s \in \mathcal{S}$, the denotation $[\![s]\!](x) \in \mathcal{Y}$ is a function mapping an input $x \in \mathcal{X}$ to an output.
    \item \textbf{Formal Grammars:} For a grammar $s \in \mathcal{S}$, the denotation $[\![s]\!](x) = L_x(s) \subseteq \Sigma_x^*$ is a language generated by the grammar under a vocabulary choice $x \in \{1, \dots, 5\}$, where $\Sigma_x$ is the terminal alphabet induced by that vocabulary.
\end{enumerate}

Let $M_\theta$ be a language model with parameters $\theta$. We distinguish:
\begin{itemize}
    \item a context $c_s$, which indicates which acts as a pointer to an instruction $s$ (e.g., the problem statement, the task name, or a grammar identifier), and
    \item an input $x$, which is the instance argument to that instruction.
\end{itemize}
In the algorithmic execution domain, $x$ is the concrete program input. In the grammar domain, $x$ is the vocabulary choice (one of five options).

An idealised model behaves like a universal interpreter, able to execute any instruction given in-context: $M_\theta(s, x) \approx [\![s]\!](x)$. PBB asks whether \emph{training} can instead `compile' such an instruction into the model's parameters $\theta$ so that only a context $c_s$ acting as a pointer to the instruction is needed at inference time.

We now define the two training losses used throughout:
\begin{itemize}
    \item \textbf{Declarative loss} (instruction modelling): 
    \[
    \mathcal{L}_\text{decl}(\theta; c_s, s) := -\log p_\theta(s | c_s), 
    \]
    the next-token negative log-likelihood of the instruction text given a task context $c_s$.
    \item \textbf{Execution loss} (behaviour modelling): 
    \[
    \mathcal{L}_\text{exec}(\theta; c_s, x, y) := -\log p_\theta(y | c_s, x),
    \]
    the next-token loss on a solution $y$ given a task context $c_s$ and the input $x$. When demonstrations are provided as execution traces, optimising for the the execution loss alone corresponds to algorithm distillation \citep{Laskin23, gandhi24}. In RL variants for settings where the model can output intermediate chain-of-thought reasoning tokens, $\mathcal{L}_\text{exec}$ is replaced by the corresponding RL objective. 
\end{itemize}

Partial evaluation \citep{Jones1993Partial} is the classical process of specialising a general interpreter to a fixed program, yielding a residual program that no longer needs the original instructions as input. Analogously, we view PBB as applying a learned specialisation mapping $\Phi$ through gradient-based training: $\theta' = \Phi(\theta; \mathcal{L})$, where $\mathcal{L}$ is the training objective used to induce specialisation.

\subsection{PBB Training Curricula}

Our experiments reveal two curriculum regimes that elicit PBB: \textbf{proactive} and \textbf{retroactive}.

Let the instruction set partition as $\mathcal{S} = \mathcal{S}_\text{train} \cup \mathcal{S}_\text{eval}$, corresponding to instructions for train and test behaviours respectively. For each train instruction $s_i \in \mathcal{S}_\text{train}$, we assume a set of execution examples $\mathcal{D}_i = \{(c_{ij}, x_{ij}, y_{ij})\}$, and write $\mathcal{D}_\text{train} = \{(s_i, \mathcal{D}_i)\}_i$.

\subsubsection*{Proactive PBB}
In proactive PBB, Stage 1 trains a general correspondence between instruction representations and behaviour using the train set, and Stage 2 then internalises evaluation instructions using only declarative supervision. 

\paragraph{Stage 1 (train set; mixed objective).} Stage 1 optimises a mixture of the declarative and execution losses on the train instructions:
\[
\begin{aligned}
\min_\theta \big[
\mathbb{E}_{s \sim \mathcal{S}_\text{train}}
\big[\mathcal{L}_\text{decl}(\theta; s)\big] +
\mathbb{E}_{\mathcal{D} \sim \mathcal{D}_\text{train}}
\mathbb{E}_{(c, x, y) \sim \mathcal{D}}
\big[\mathcal{L}_\text{exec}(\theta; c, x, y)\big]\big].
\end{aligned}
\]

Concretely, this can be implemented by interleaving the two formats of training examples.

\paragraph{Stage 2 (evaluation instructions).} Stage 2 exposes the model to evaluation instructions $s \in \mathcal{S}_\text{eval}$ without executions, training only on $\mathcal{L}_\text{decl}$. In this regime, the specialisation operator is the gradient update induced by instruction modelling:

\[
\Phi_\text{pro}(\theta, \mathcal{L}_\text{decl}) := \theta - \eta \nabla_\theta \mathcal{L}_\text{decl}(\theta; c_s, s).
\]

Stage 1 is what makes this update behave like `compilation': because $\theta$ has been trained so that gradients through $s$ are predictive of the execution-relevant updates, training on $s$ alone can internalise $[\![s]\!]$ into the weights.

\subsubsection*{Retroactive PBB}

In retroactive PBB, Stage 1 learns all instructions declaratively (without execution competence), and Stage 2 introduces execution supervision for only $\mathcal{S}_\text{train}$.

\paragraph{Stage 1 (all instructions).} Train on all instructions:
\[
\min_\theta \mathbb{E}_{s \sim \mathcal{S}}[\mathcal{L}_\text{decl}(\theta; s)].
\]

\paragraph{Stage 2 (train executions).} The retroactive specialisation operator is:
\[
\Phi_\text{ret}(\theta, \mathcal{L}_\text{exec}) := \theta - \eta\sum_{(c_s, x, y) \in \mathcal{D}_\text{train}} \nabla_\theta \mathcal{L}_\text{exec}(\theta; c_s, x, y).
\]

Intuitively, although Stage 2 gradients are computed only from executions of paired instructions, they update shared parameters $\theta$. This global update can `activate' the latent declarative knowledge acquired in Stage 1, facilitating execution at test time even for evaluation instructions that never received direct execution supervision.

\section{Experimental Setup}
\label{setup}

To investigate Programming by Backprop (PBB), we conduct experiments across two distinct domains: algorithmic execution of Python code and sentence generation under formal grammars. 

\subsection{Datasets and Tasks}

We create three synthetic datasets to provide a controlled environment for studying PBB. We additionally experiment with a real-world coding dataset and corresponding execution task. For each dataset, we define a set of instructions $\mathcal{S}$, which is partitioned into a train and evaluation set.

\paragraph{Algorithmic Execution.} These tasks test whether models can learn to execute functions from their source code. 

\begin{itemize}
    \item \textbf{Random Arithmetic:} This dataset contains 1,000 unique Python functions that map integers to integers. The functions are synthetically generated by composing basic control flow (\texttt{for} loops, \texttt{if} / \texttt{else} conditionals) and arithmetic operators (\texttt{+}, \texttt{-}, \texttt{*}, \texttt{//}, \texttt{\%}, \texttt{>}, \texttt{<}, \texttt{exp}, \texttt{abs}). This allows us to control for procedural complexity (i.e., the number of operations). For our main experiments, we use 100 functions for $\mathcal{S}_\text{train}$ and 100 for $\mathcal{S}_\text{eval}$. With this dataset, we also define compositions of two functions to evaluate models' ability to compose behaviours defined by instructions that were trained on independently. We give these compositional functions unique names and define the instructions as, for example, \texttt{def Blorp(x): Zibble(Snurg(x))}. Execution supervision is also included for composite train instructions.
    \item \textbf{Leetcode:} This dataset consists of 702 real-world algorithmic problems and their Python solutions, sourced from the competitive programming platform. This tests PBB on more complex and naturalistic programs. We use 500 problems for $\mathcal{S}_\text{train}$ and 100 for $\mathcal{S}_\text{eval}$. 
    \item \textbf{Ciphers:} To test generalisation to novel, OOD programs, we create three custom ciphers (Alice, Bob, Kevin) that are variations of standard ciphers (Caesar, Atbash, Vigenère). We assume these novel ciphers are absent from the model's pretraining data, allowing for controlled experimentation. The same 500 Leetcode problems are used for $\mathcal{S}_\text{train}$ and the three custom ciphers form $\mathcal{S}_\text{eval}$. For this task, we also generate execution demonstrations from an imbalanced distribution (Appendix~\ref{ap:ciphers}), reflecting the greater occurrence of examples with specific shift values (e.g., ROT13) in web data \citep{mccoy2024embers}. This allows us to compare learning ciphers in an input-general fashion via PBB to learning from skewed, input-specific demonstration data. 
\end{itemize}

For all tasks, input-output examples are framed as word problems. We generate ground-truth solutions by executing the corresponding Python code. To test the benefits of intermediate reasoning, we also generate chain-of-thought (CoT) solutions for each problem using \texttt{GPT-4o} in a post-rationalisation step \citep{Zelikman2022Star}. We also use \texttt{GPT-4o} to generate semantically equivalent natural language instructions for the Random Arithmetic Python functions, allowing us to evaluation the impact of algorithm representation on the performance of PBB.






\paragraph{Generation Under Formal Grammars.} 
To test PBB beyond code, we construct a procedurally generated suite of artificial grammars that define syntactic constraints on sentence formation. Concretely, we generate a set of 200 unique context-free grammars (CFGs). Each grammar is produced by sampling from a compact, interpretable parameter space that controls typological properties (word order families such as SVO, SOV, VSO, etc.), modifier placement (adjectives pre/post-nominal, determiners pre/post-nominal, adverbs pre/post-verbal), and optional structural features. Each grammar is represented symbolically as a small set of production rules (nonterminals and productions). We use five vocabularies (\texttt{Vocab-A}, ..., \texttt{Vocab-E}), each containing ten distinct nouns, verbs, and adjectives shared across all grammars for sampling strings.

We sample derivation trees from each CFG by repeatedly expanding nonterminals according to that grammar until a depth cutoff. A sampled derivation tree is used to produce a sentence by instantiating terminals from the specified vocabulary. From the 200 grammars, we designate 100 as $\mathcal{S}_\text{train}$ (grammars paired with example strings) and 100 as $\mathcal{S}_\text{eval}$ (grammars for which symbolic specification is shown during training, but not example strings). We evaluate model generations with a strict, grammar-based validity test: a candidate sentence is accepted if it can be parsed by the requested CFG (we use a chart/Earley parser). Accuracy on a grammar is thus the fraction of model outputs that produce at least one valid parse under that grammar. For this task, CoT is not used during training or evaluation.

\subsection{Data Augmentation}

Empirically, we find that it is important to augment $c_s$ when training on instructions for PBB to work effectively (\S\ref{sec:algorithmic_results}). This is in line with prior work on out-of-context reasoning \citep{berglund2023oocr}. We therefore define 20 context templates for each dataset that act as pointers to instructions (e.g., for Leetcode: `Write a Python function that solves the following problem $\{s\}$', `I need a Python implementation that does $\{s\}$', ...). A single context template is sampled uniformly at random for each instruction. 30 unique execution examples are provided per train instruction during execution supervision.

\subsection{Training Details}

We use instruction-tuned \texttt{Qwen3} models (4B, 8B, 14B) \citep{qwen2025qwen25technicalreport} as our primary set of base models and conduct additional experiments with \texttt{GPT-4o} \citep{openai2024gpt4ocard} via the OpenAI finetuning API to investigate PBB in a large frontier model. We also repeat core experiments with instruction-tuned \texttt{Llama-3} models (1B, 3B, 8B) \citep{llama3} to see if trends are consistent across model families (Appendix~\ref{ap:llama}). All training stages use five epochs, meaning a single instruction or execution example is encountered five times in total. For RL runs, we use GRPO \citep{shao2024deepseekmathpushinglimitsmathematical} with a group size of 8 and no KL regularisation ($\beta = 0$). A positive reward ($+1$) is given if the final answer is properly formatted and correct; a neutral reward ($0$) is given if the answer is properly formatted but incorrect; a negative reward ($-1$) is given if the model fails to produce a final answer in the required format. For all training runs, the batch size is set to 32 and we use a constant learning rate of $1 \times 10^{-5}$. We use a sampling temperature $t = 0.8$ for evaluation and RL training. All evaluations are averaged over 16 samples and 95\% confidence intervals are reported.

\section{Results}

\subsection{PBB for Algorithmic Execution}
\label{sec:algorithmic_results}

\begin{figure*}[t]
\begin{center}
\includegraphics[width=0.98\linewidth]{"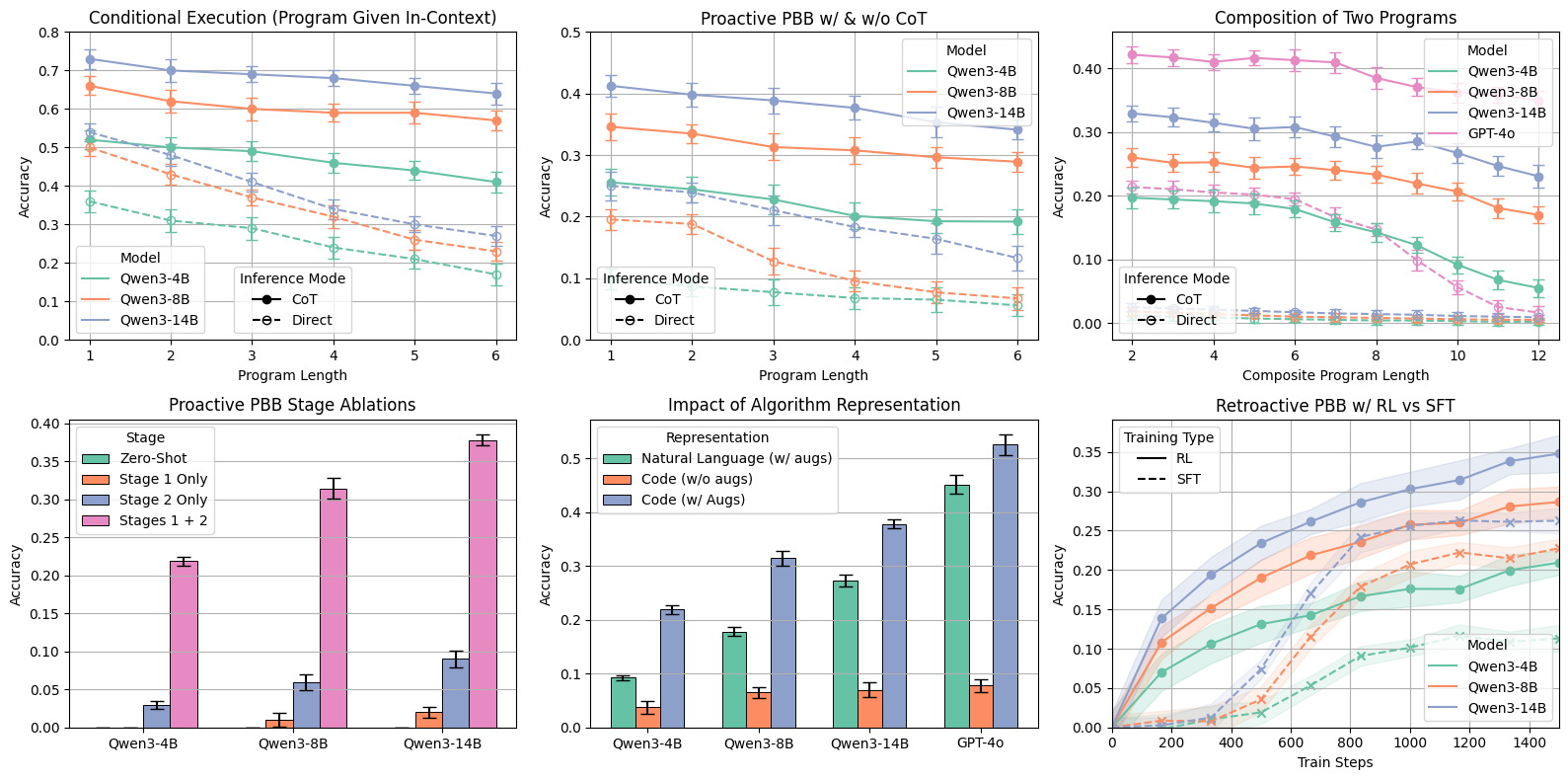"}
\end{center}
\vspace{-1em}
\caption{\textbf{Upper Left}: Accuracy when conditionally executing random arithmetic programs of different lengths provided in-context. \textbf{Upper Middle}: Accuracy following proactive PBB on executing unpaired programs. Explicitly computing intermediate operations via CoT is more effective than direct inference, but larger models show some ability to encode entire procedures in their weights via PBB. \textbf{Upper Right}: Accuracy following retroactive PBB throughout stage 2 training. RL is more effective, with SFT initially memorising paired executions before generalising execution ability to unpaired programs. \textbf{Lower Left:} Accuracy following each stage of proactive PBB. The full curriculum has a significant impact on performance. \textbf{Lower Middle:} Accuracy following proactive PBB when programs are represented as natural language or code. The role of prompt augmentations is also ablated. \textbf{Lower Right:} Accuracy following proactive PBB for compositions of two programs that have been trained on independently. Direct inference performance is only shown for \texttt{GPT-4o} as the smaller models fail in this setting.}
\label{fig:random}
\vspace{-1em}
\end{figure*}

We first evaluate whether models can learn to execute algorithms from instructions in training data. Our experiments on the Random Arithmetic and Leetcode domains yield several key findings regarding the conditions required for PBB and the nature of the resulting behaviour.

\paragraph{Role of curricula.}
We find that simply including instructions in the training data is insufficient for models to acquire executable procedural knowledge. As shown in the stage ablation (Fig.~\ref{fig:random}, lower left), training on stage 1 train instructions and executions, or stage 2 evaluation instructions in isolation results in negligible-to-low performance on test behaviour. The zero-shot performance is zero for all models because we are working with synthetically generated Python programs that models would have no prior knowledge of. Under the full `proactive' curriculum, where the model first learns the correspondence between instructions and behaviour on the train set (stage 1) before internalising evaluation instructions (stage 2), \texttt{Qwen3-14B} reaches 37\% average execution accuracy for test behaviours. 

\paragraph{Reliability and implicit execution.}
While PBB induces executable behaviour, it remains less reliable than explicitly conditioning the model on the instruction in-context (Fig.~\ref{fig:random}, upper left). In spite of this, PBB enables some degree of direct inference: models can produce the correct output without generating intermediate chain-of-thought (CoT) reasoning steps, albeit with lower accuracy than when CoT is used (Fig.~\ref{fig:random}, upper middle). This suggests the model is capable of performing the necessary multi-step algorithmic operations implicitly within the forward pass, even without the instruction being in-context. Furthermore, we observe that models can compose independently learned instructions: Fig.~\ref{fig:random} (upper right) shows that models can execute composites of two programs they were trained on separately, with larger models like \texttt{GPT-4o} showing some capability even without explicit reasoning traces.

\paragraph{Retroactive PBB: RL vs. SFT.}
In the retroactive regime, where declarative knowledge is `activated' by later train set execution supervision, we find that reinforcement learning (RL) is significantly more effective than supervised finetuning (SFT) for stage 2. As illustrated in Fig.~\ref{fig:random} (lower right), SFT tends to memorise the train execution demonstrations initially, taking more samples to generalise execution capabilities to test behaviours, whereas RL drives faster and more robust generalisation.

\paragraph{Impact of algorithm representation.}
The format of the instruction significantly influences PBB performance. We compare instructions provided as Python source code against semantically equivalent natural language descriptions. Results indicate that algorithms are learned more effectively when expressed in code (Fig.~\ref{fig:random}, lower middle), although the performance gap is lower for larger models. This gap may stem from the potential ambiguity and verbosity of natural language, or arguably because current LLMs possess a stronger inductive bias for reasoning from code due to their pretraining distributions. Augmentations applied to the prompts for the instructions in training data are found to be essential to the outcome of PBB.

\begin{figure*}[t]
\begin{center}
\includegraphics[width=0.8\linewidth]{"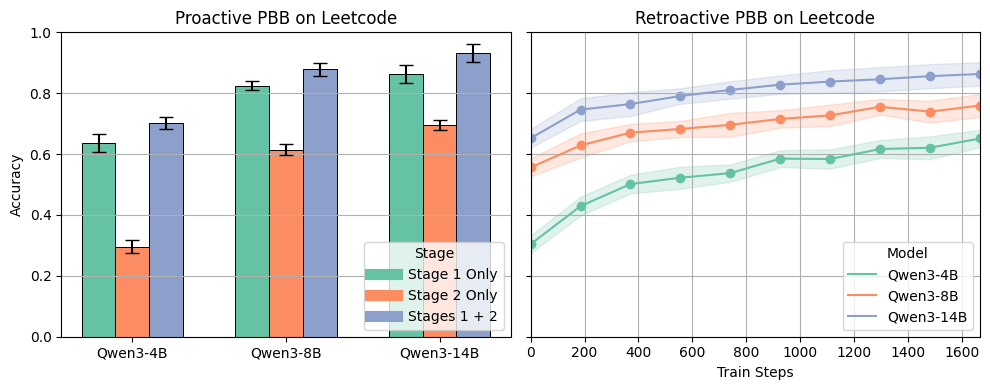"}
\end{center}
\vspace{-1em}
\caption{\textbf{Left:} Accuracy following each stage of proactive PBB on Leetcode programs. Despite zero-shot performance being high in this setting, the full curriculum yields the greatest performance, showing that execution performance can be refined by further training on instructions. \textbf{Right:} Accuracy following retroactive PBB (RL) throughout Stage 2 training.}
\label{fig:leetcode}
\vspace{-1em}
\end{figure*}

\paragraph{PBB on pretrained domains (Leetcode).}
We test PBB on the Leetcode dataset to assess its utility for tasks where the model has prior exposure to similar programs during pretraining. Even here, the full proactive curriculum yields the highest performance on test behaviours (Fig.~\ref{fig:leetcode}). In contrast, retroactive PBB performs comparably to simply training on the train executions, suggesting that the retroactive activation of latent knowledge is less beneficial when the domain is already familiar.

\subsection{Sample Efficient Learning via PBB} \label{subsec:results:sampleefficiency}

A motivation for PBB is the potential for data efficiency --- replacing many execution examples with a single high-level instruction.

\begin{wrapfigure}{r}{0.49\textwidth}
\vspace{-2em}
\centering
\includegraphics[width=0.45\textwidth]{"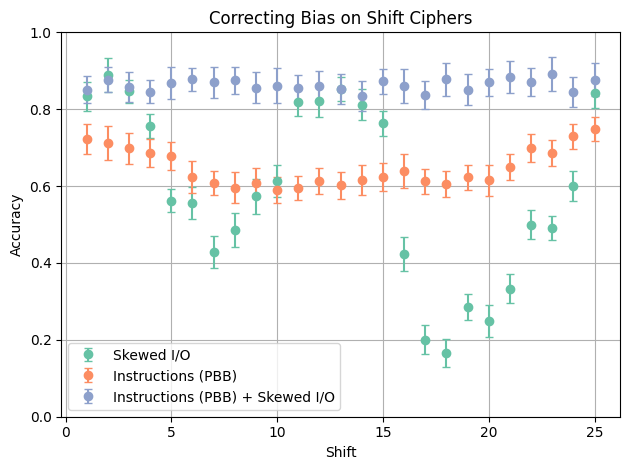"}
\vspace{-1em}
\caption{Finetuning \texttt{GPT-4o} on custom Ciphers via proactive PBB (Stage 1 Leetcode, Stage 2 Ciphers) yields greater robustness to shift variations than training on imbalanced demonstrations. Mixing Stage 2 training on cipher code with imbalanced demonstrations yields the best results, overcoming biases from demonstrations while grounding execution behaviour.}
\label{fig:efficiency}
\end{wrapfigure}

\paragraph{High sample efficiency.}
We compare PBB against algorithm distillation (AD) \citep{Laskin23, gandhi24}, where the model learns to implement an algorithm from input-specific examples. On the Random Arithmetic task, we find that PBB is highly sample efficient. As shown in Fig.~\ref{fig:random_efficiency}, training on a single instruction in stage 2 of the proactive curriculum yields execution fidelity comparable to, or better than, training on up to 100 execution traces (for CoT inference) or input-output pairs (for direct inference). To ensure fair comparison, algorithm distillation also follows stage 1 of proactive PBB. This confirms that a concise instruction can substitute for a substantial volume of demonstration data. However, we note that proactive PBB requires the amortised cost of stage 1, which primes the model for learning from instructions. Proactive PBB can therefore be seen as an approach to sample efficient generalisation given an initial meta-train phase.

\paragraph{OOD generalisation and bias correction.}
PBB also facilitates robust generalisation. We test this on the Ciphers task using previously unseen custom shift ciphers. Training on demonstration data (i.e., AD) that is skewed (e.g., an imbalance favouring specific shift values, such as ROT13, common in web data \citep{mccoy2024embers}) leads to biased performance that degrades when the shift parameter varies. In contrast, proactive PBB (stage 1 on Leetcode, stage 2 on cipher code) learns the input-general function, resulting in performance that is robust to parameter variations (Fig.~\ref{fig:efficiency}, right). In both cases, low and high shift values also have greater performance, because these lead to letters that are alphabetically close to the original and thus are less prone to errors. Notably, combining PBB with skewed demonstrations yields the best performance: the instruction mitigates the bias from the skewed data, while the demonstrations help ground the execution behaviour.

\begin{figure*}[t]
\begin{center}
\includegraphics[width=0.8\linewidth]{"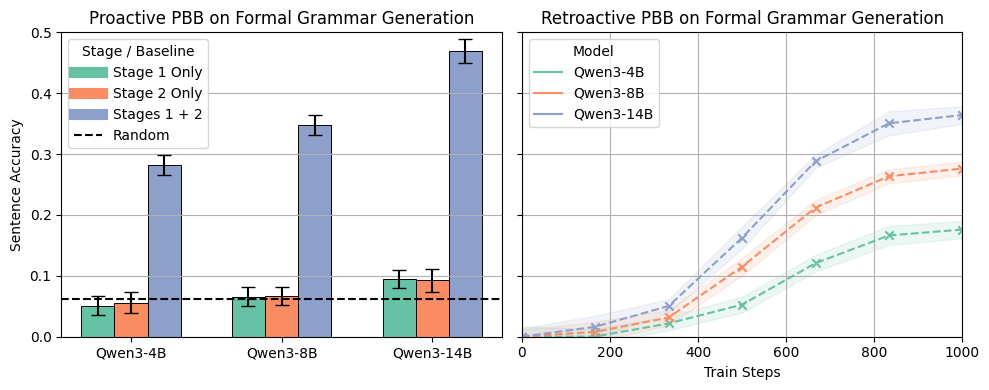"}
\end{center}
\vspace{-1em}
\caption{\textbf{Left:} Compliance to unpaired grammars following each stage of proactive PBB. Stage 1 or stage 2 alone each does comparably to randomly generating word sequences from the vocabulary. The full proactive PBB pipeline enables compliance with grammars described symbolically in training data. \textbf{Right:} Compliance to unpaired grammars following retroactive PBB (SFT).}
\label{fig:grammars}
\vspace{-1em}
\end{figure*}

\subsection{PBB for Grammar Compliance}

To demonstrate PBB beyond code execution, we evaluate its application to text generation under formal constraints.

\paragraph{Implicit grammar computation.}
We find that PBB can successfully `program' models to generate text compliant with novel context-free grammars defined in the training data. As shown in Fig.~\ref{fig:grammars}, the full proactive curriculum enables models to generate valid sentences for test grammars with accuracy far exceeding random generation from the defined vocabulary or single-stage baselines. Crucially, in this setting, the model produces compliant text immediately upon prompting, without CoT reasoning. This indicates that the model has encoded the grammar rules into its weights and is performing the necessary syntactic computations, tracking state and ensuring rule compliance, implicitly within its forward pass. Retroactive PBB also enables models to generate text that is compliant with test grammars, though at a lower fidelity than with the proactive curriculum.

\section{Conclusion}

This work shows that language models can internalize executable procedural knowledge from declarative instructions in training data via Programming by Backprop (PBB). We find that PBB emerges under structured, multi-stage curricula, is highly sample efficient, and generalises across domains. While execution remains less reliable than explicit prompting, the results demonstrate that instructions can be implicitly ``compiled'' into model parameters. This has fundamental implications for how researchers and practitioners can teach models new behaviours and represents a key finding for safety: instructions present in training data could elicit unintended behaviours at inference time, underscoring the importance of careful data curation and filtering.

\section{Reproducibility Statement}

We are open-sourcing all code and datasets needed to reproduce our experiments at \url{https://github.com/jonathan-cook235/Programming-by-Backprop}. This includes data generation scripts and training code.

\bibliography{iclr2026_conference}
\bibliographystyle{iclr2026_conference}

\newpage
\appendix

\section{Llama Results}
\label{ap:llama}

\begin{figure}[h]
\begin{center}
\includegraphics[width=0.9\linewidth]{"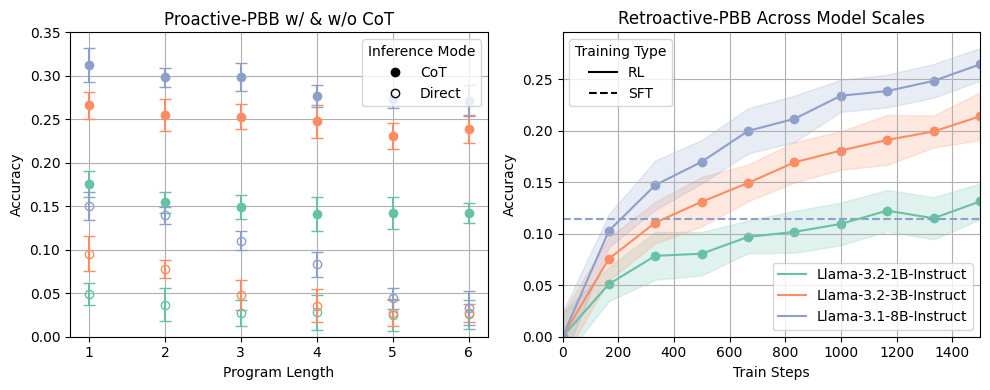"}
\end{center}
\caption{Results with \texttt{Llama-3} models on Random Arithmetic.}
\label{fig:llama_random}
\end{figure}

\begin{figure}[h]
\begin{center}
\includegraphics[width=0.9\linewidth]{"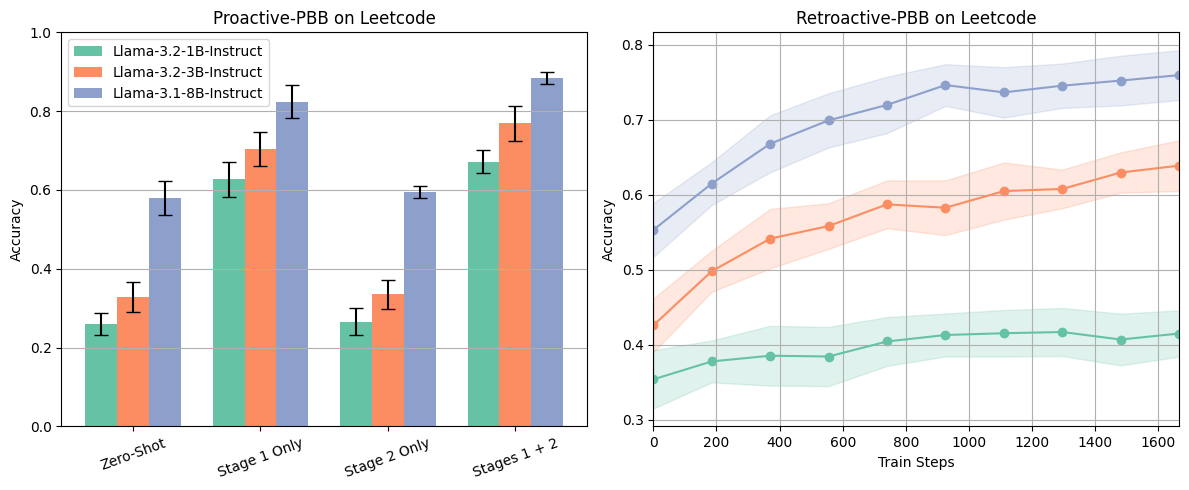"}
\end{center}
\caption{Results with \texttt{Llama-3} models on Leetcode.}
\label{fig:llama_leetcode}
\end{figure}

\begin{figure}[h]
\begin{center}
\includegraphics[width=0.5\linewidth]{"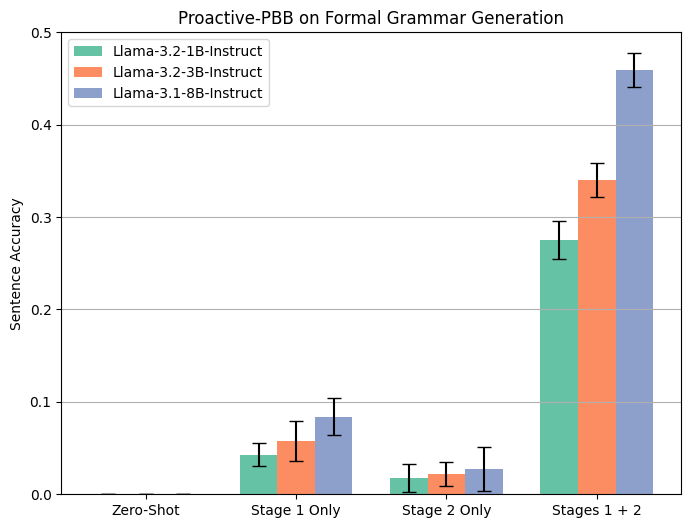"}
\end{center}
\caption{Proactive PBB with \texttt{Llama-3} models for generation under formal grammars.}
\label{fig:llama_grammars}
\end{figure}

\newpage
\section{Data Scaling}
\label{ap:data}

\subsection{Ablation over Dataset Size}
\label{app:abl}
\begin{figure}[h!]
\begin{center}
\includegraphics[width=0.48\linewidth]{"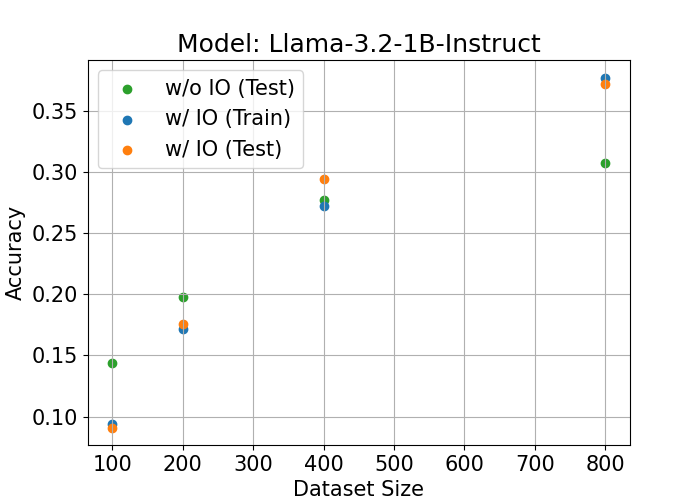"}
\includegraphics[width=0.48\linewidth]{"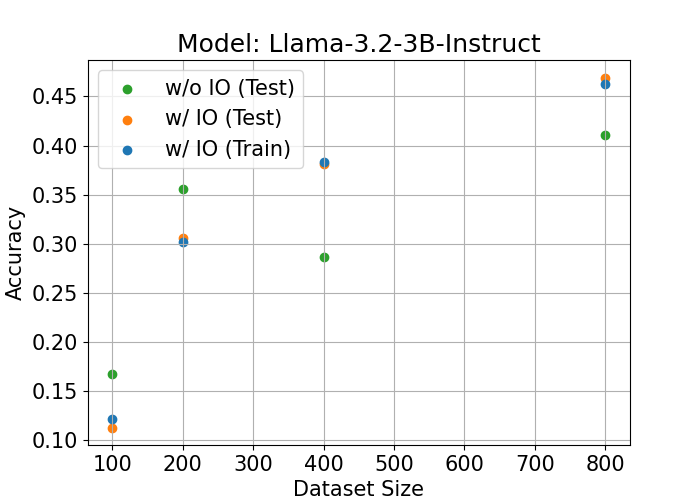"}
\includegraphics[width=0.48\linewidth]{"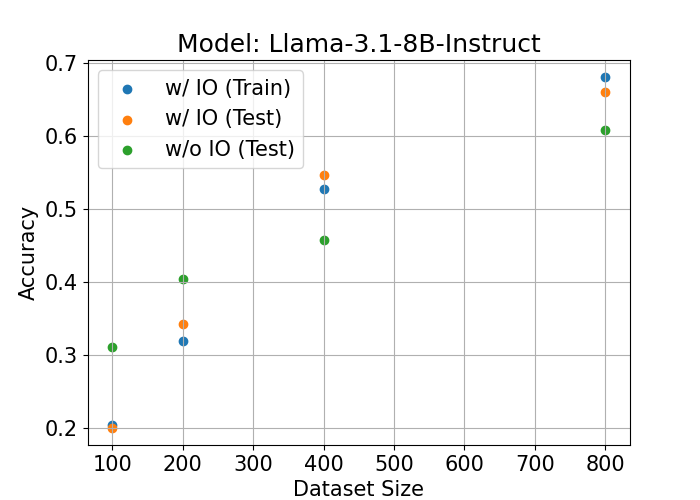"}
\end{center}
\caption{Performance comparison of Llama models across 1B, 3B and 8B on paired (\textit{w/ IO}) and unpaired (\textit{w/o IO}) Random Arithmetic program evaluation. Each model is trained and tested across varying dataset sizes. Dataset size refers to the number of unique functions present in the dataset.}
\label{fig:data_scaling}
\end{figure}

Figure \ref{fig:data_scaling} compares the performance of Llama models (1B, 3B and 8B parameters) for varying dataset size on the evaluation of Random Arithmetic programs. Here, `dataset size', refers specifically to the amount of unique code functions included in the dataset. Performance is evaluated on three separate sets:
\begin{itemize}
    \item The \textit{w/ IO Train} set: both the function and the IO pairs are observed during training
    \item The \textit{w/ IO Test} set: uses the same functions as \textit{w/ IO Train} but different IO pairs, not included in the training data
    \item The \textit{w/o IO Test} set: evaluates IO pairs for functions seen only as code during training
\end{itemize}
The results show that accuracy on both \textit{w/ IO} and \textit{w/o IO} sets generally increases with larger dataset sizes and larger model scales. Notably, model performance is strongly tied to parameter count; for example, the 8B model trained on only 100 unique functions achieves comparable performance on the \textit{w/o IO} set to the 1B model trained on 800 functions. The number of unpaired (\emph{w/o IO}) functions is fixed at 200. The upper limit of 800 paired (\emph{w/ IO}) functions reflects the total of 1000 available functions that we generated for Random Arithmetic.

\begin{figure}[h!]
\begin{center}
\includegraphics[width=0.5\linewidth]{"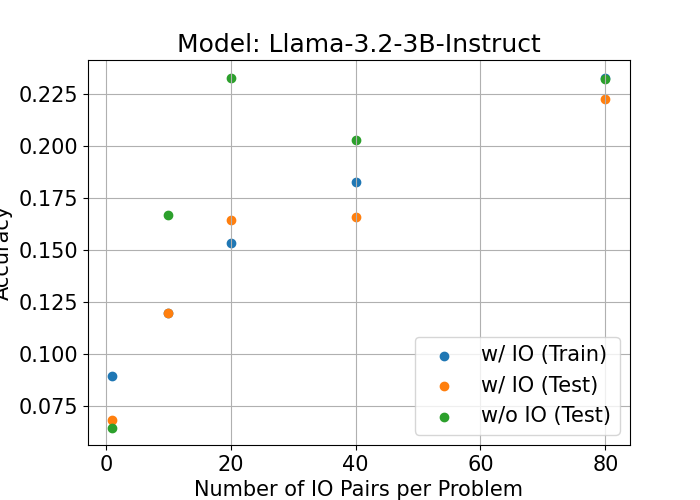"}
\end{center}
\caption{Impact of varying the number of IO training pairs for paired (\textit{w/ IO}) programs and unpaired (\textit{w/o IO}) sets evaluation accuracy. Results are shown for the \texttt{Llama-3.2-3B-Instruct} model using a Random Arithmetic dataset comprising 200 distinct functions.}
\label{fig:io_scaling}
\end{figure}

\subsection{Ablation over Number of IO Pairs}
In Figure \ref{fig:io_scaling} we vary the number of IO training pairs (per program) provided for the \textit{w/ IO} set, and examine the results.
This analysis specifically uses the \texttt{Llama-3.2-3B-Instruct} model on the Random Arithmetic dataset, which for this experiment consists of 200 distinct functions. Performance is reported across the same sets as the ones described in Appendix \ref{app:abl}. The results show how increasing the quantity of IO examples for each program affects not only direct generalisation in the \textit{w/ IO} Test set, but also the model's ability to accurately execute \textit{w/o IO} programs.

\newpage
\section{Single-Stage Programming by Backprop}
\label{ap:single-stage}

\begin{figure}[h!]
\begin{center}
\includegraphics[width=0.5\linewidth]{"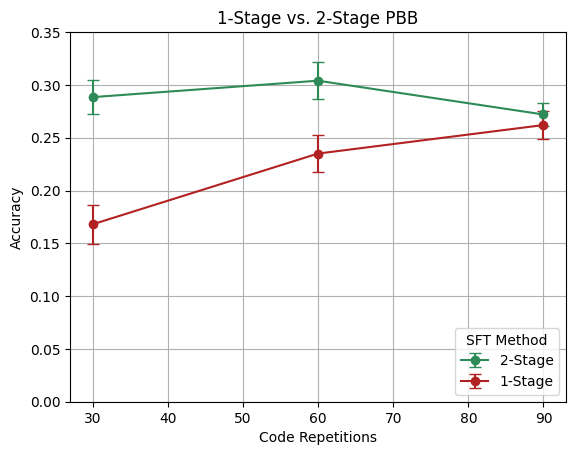"}
\end{center}
\caption{Comparing two-stage proactive PBB to a single SFT stage on the full Random Arithmetic training data mixture for different numbers of repeated source code samples. The base model is Llama-3.1-8B-Instruct.}
\label{fig:single-stage}
\end{figure}

In Figure \ref{fig:single-stage}, we show the accuracy of Llama-3.1-8B-Instruct on unpaired (\emph{w/o IO}) Random Arithmetic program evaluation following proactive PBB in comparison to a single SFT stage with all training data in a single mixture. As we scale the number of times the same piece of unpaired (\emph{w/o IO}) source code appears in the dataset, with prompt and response preamble augmentations, single-stage SFT approaches the performance of proactive PBB. The greater sample efficiency of proactive PBB is likely because initial train steps on source code are waisted in single-stage SFT, as a code-I/O relationship has not yet been learned. 

\newpage
\section{Online vs. Offline Retroactive-PBB}
\label{ap:online}

\begin{figure}[h!]
\begin{center}
\includegraphics[width=0.5\linewidth]{"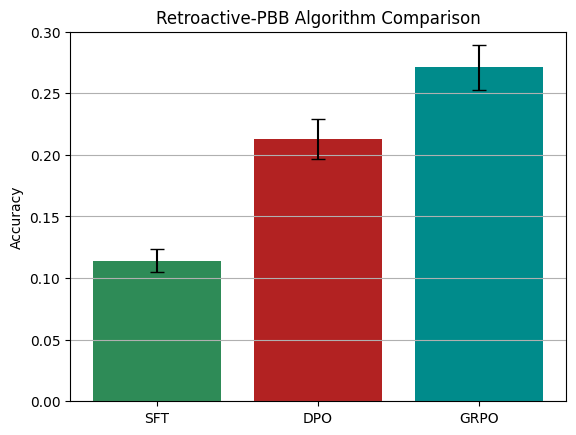"}
\end{center}
\caption{Comparing finetuning algorithms for the second stage of retroactive PBB on Random Arithmetic with Llama-3.1-8B-Instruct. DPO is an offline method, but allows for learning from positive and negative examples. GRPO is online and thus has the added benefit of learning from on-policy data.}
\label{fig:dpo}
\end{figure}

In Figure \ref{fig:dpo}, we compare different finetuning algorithms for the second stage of retroactive PBB with Llama-3.1-8B-Instruct on Random Arithmetic. DPO allows for learning from both positive and negative samples, considerably outperforming SFT. GRPO is an online RL algorithm, meaning that the model learns from on-policy data, which could be why it yields further improvements.

\newpage
\section{Ciphers Data}
\label{ap:ciphers}

\begin{figure}[h!]
\begin{center}
\includegraphics[width=0.5\linewidth]{"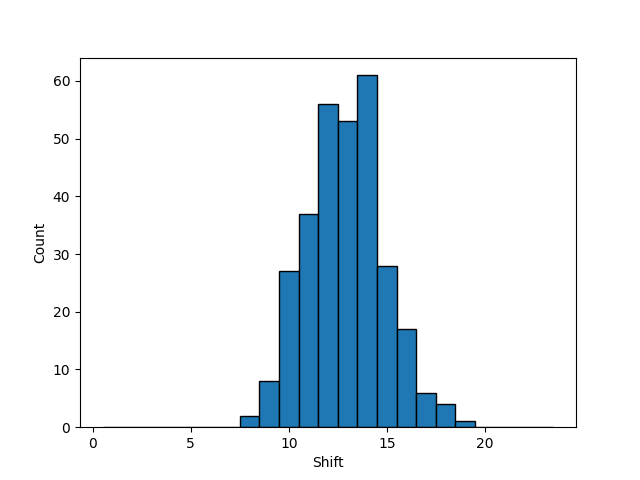"}
\end{center}
\caption{Sampled shifts for cipher I/O pairs.}
\label{fig:shifts}
\end{figure}

A plot showing the distribution of IO pairs used in Figure \ref{fig:efficiency} is provided in Figure \ref{fig:shifts}.

\newpage
\section{Natural Language Descriptions}
\label{ap:example}

Here, we include an example of a random arithmetic program and its natural language description.

\textbf{Program:} \begin{verbatim}
    def Blaankle(x):
       t0 = x + x
       t1 = 1 * abs(t0)
       return t1
\end{verbatim}

\textbf{Description:} \textit{A Blaankle is a process that takes an input value, doubles it, and then returns the absolute value of the doubled result.}

\subsection{Discussion on Natural Language PBB}

Current models exhibit a strong dependence on the structure of the description. Code and grammars provide explicit, unambiguous abstractions that models can internalise more reliably. Natural language (NL) lacks this regular structure and therefore poses a more challenging learning signal for PBB.

However, we see this as a scaling and representation issue rather than a fundamental limitation:

\begin{itemize}
    \item \textbf{Positive scale-dependence:} NL PBB improves with model size, suggesting that future models may be substantially more capable in this regime.
    \item \textbf{Intermediate formalisms:} Many practical workflows already translate NL specifications to structured representations (pseudocode, planning languages). PBB could be applied at these intermediate stages; our work provides evidence that the formal end of this spectrum works well.
    \item \textbf{Synthetic datasets at scale:} Because code-like descriptions are easy to generate programmatically, PBB could be elicited in earlier LLM training stages, potentially allowing models to internalise general symbolic-interpretation skills before encountering downstream NL descriptions.
    \item \textbf{General mechanisms:} Our CFG experiments demonstrate that PBB extends outside programming entirely, to abstract formal systems. This indicates that what matters is symbolic structure, not code specifically.
\end{itemize}

Thus, while NL PBB is currently weaker, the paradigm itself is not restricted to code, and our results point towards concrete avenues for future work on making NL PBB practical.

\newpage
\section{Compute Requirements}

All experiments with \texttt{Llama} models can be run on two GPUs with 40GB vRAM. We used data parallelism over 4 NVIDIA L40s GPUs to run these experiments. 

Experiments with \texttt{GPT-4o} made use of the OpenAI finetuning API. Data generation (Leetcode word problems and post-rationalised chain-of-thought ground truth outputs for all datasets) and finetuning runs came to a total cost just over 500 USD. 

\newpage
\section{Additional Ablations}

\begin{figure}[h]
\begin{center}
\includegraphics[width=0.9\linewidth]{"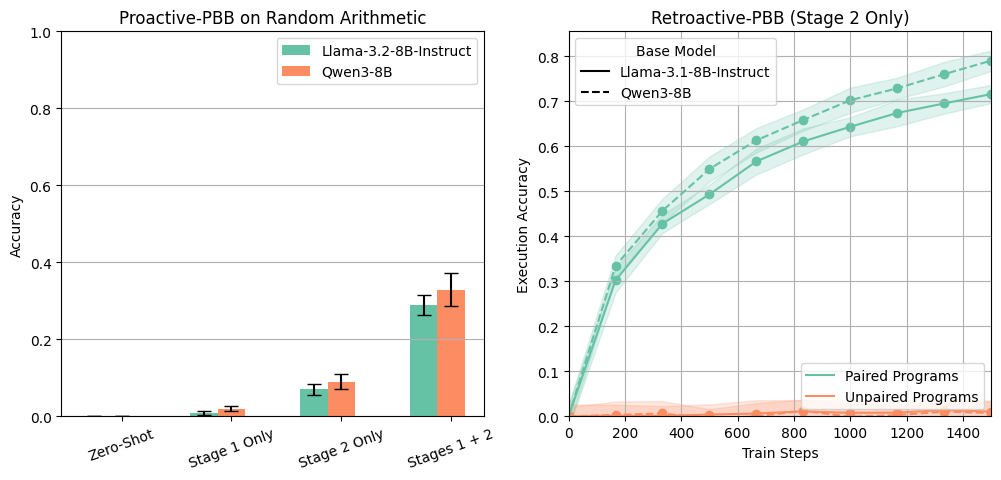"}
\end{center}
\caption{\textbf{Left}: Accuracy following each stage of proactive PBB on executing unpaired Random Arithmetic programs. \textbf{Right}: Accuracy for test inputs following only stage 2 (``activation'') of retroactive PBB on paired vs. unpaired Random Arithmetic programs. This training corresponds to only doing RL on execution problems with train inputs on paired programs.}
\label{fig:qwen_random}
\end{figure}

\end{document}

%% file: math_commands.tex
\usepackage{amsmath,amsfonts,bm}

\def\eqref#1{equation~\ref{#1}}

\def\1{\bm{1}}

\DeclareMathAlphabet{\mathsfit}{\encodingdefault}{\sfdefault}{m}{sl}
\SetMathAlphabet{\mathsfit}{bold}{\encodingdefault}{\sfdefault}{bx}{n}



%% file: iclr2026_conference.bib
@misc{aryabumi2024code,
      title={To Code, or Not To Code? Exploring Impact of Code in Pre-training}, 
      author={Viraat Aryabumi and Yixuan Su and Raymond Ma and Adrien Morisot and Ivan Zhang and Acyr Locatelli and Marzieh Fadaee and Ahmet Üstün and Sara Hooker},
      year={2024},
      eprint={2408.10914},
      archivePrefix={arXiv},
      primaryClass={cs.CL},
      url={https://arxiv.org/abs/2408.10914}, 
}

@inproceedings{
ruis2025procedural,
title={Procedural Knowledge in Pretraining Drives Reasoning in Large Language Models},
author={Laura Ruis and Maximilian Mozes and Juhan Bae and Siddhartha Rao Kamalakara and Dwaraknath Gnaneshwar and Acyr Locatelli and Robert Kirk and Tim Rockt{\"a}schel and Edward Grefenstette and Max Bartolo},
booktitle={The Thirteenth International Conference on Learning Representations},
year={2025},
url={https://openreview.net/forum?id=1hQKHHUsMx}
}

@inproceedings{
piterbarg2025training,
title={Training Language Models on Synthetic Edit Sequences Improves Code Synthesis},
author={Ulyana Piterbarg and Lerrel Pinto and Rob Fergus},
booktitle={The Thirteenth International Conference on Learning Representations},
year={2025},
url={https://openreview.net/forum?id=AqfUa08PCH}
}

@inproceedings{
nikankin2025arithmetic,
title={Arithmetic Without Algorithms: Language Models Solve Math with a Bag of Heuristics},
author={Yaniv Nikankin and Anja Reusch and Aaron Mueller and Yonatan Belinkov},
booktitle={The Thirteenth International Conference on Learning Representations},
year={2025},
url={https://openreview.net/forum?id=O9YTt26r2P}
}

@article{berglund2023oocr,
  publtype={informal},
  author={Lukas Berglund and Asa Cooper Stickland and Mikita Balesni and Maximilian Kaufmann and Meg Tong and Tomasz Korbak and Daniel Kokotajlo and Owain Evans},
  title={Taken out of context: On measuring situational awareness in LLMs},
  year={2023},
  cdate={1672531200000},
  journal={CoRR},
  volume={abs/2309.00667},
  url={https://doi.org/10.48550/arXiv.2309.00667}
}

@inproceedings{
betley2025tell,
title={Tell me about yourself: {LLM}s are aware of their learned behaviors},
author={Jan Betley and Xuchan Bao and Mart{\'\i}n Soto and Anna Sztyber-Betley and James Chua and Owain Evans},
booktitle={The Thirteenth International Conference on Learning Representations},
year={2025},
url={https://openreview.net/forum?id=IjQ2Jtemzy}
}

@article{
chen2023program,
title={Program of Thoughts Prompting: Disentangling Computation from Reasoning for Numerical Reasoning Tasks},
author={Wenhu Chen and Xueguang Ma and Xinyi Wang and William W. Cohen},
journal={Transactions on Machine Learning Research},
issn={2835-8856},
year={2023},
url={https://openreview.net/forum?id=YfZ4ZPt8zd},
note={}
}

@article{mccoy2024embers,
author = {R. Thomas McCoy  and Shunyu Yao  and Dan Friedman  and Mathew D. Hardy  and Thomas L. Griffiths },
title = {Embers of autoregression show how large language models are shaped by the problem they are trained to solve},
journal = {Proceedings of the National Academy of Sciences},
volume = {121},
number = {41},
pages = {e2322420121},
year = {2024},
doi = {10.1073/pnas.2322420121},
URL = {https://www.pnas.org/doi/abs/10.1073/pnas.2322420121},
eprint = {https://www.pnas.org/doi/pdf/10.1073/pnas.2322420121}}

@inproceedings{Krasheninnikov24,
  author       = {Dmitrii Krasheninnikov and
                  Egor Krasheninnikov and
                  Bruno Kacper Mlodozeniec and
                  Tegan Maharaj and
                  David Krueger},
  title        = {Implicit meta-learning may lead language models to trust more reliable
                  sources},
  booktitle    = {Forty-first International Conference on Machine Learning, {ICML} 2024,
                  Vienna, Austria, July 21-27, 2024},
  publisher    = {OpenReview.net},
  year         = {2024},
  url          = {https://openreview.net/forum?id=Fzp1DRzCIN}
}

@misc{Wang24,
  author       = {Boshi Wang and
                  Xiang Yue and
                  Yu Su and
                  Huan Sun},
  title        = {Grokked Transformers are Implicit Reasoners: {A} Mechanistic Journey
                  to the Edge of Generalization},
  year         = {2024},
  url          = {https://doi.org/10.48550/arXiv.2405.15071},
  archivePrefix={arXiv},
  eprint       = {2405.15071}
}

@misc{Petty24,
  author       = {Jackson Petty and
                  Sjoerd van Steenkiste and
                  Tal Linzen},
  title        = {How Does Code Pretraining Affect Language Model Task Performance?},
  year         = {2024},
  url          = {https://doi.org/10.48550/arXiv.2409.04556},
  archivePrefix= {arXiv},
  eprint       = {2409.04556}
}

@misc{Yang24,
  author       = {Sohee Yang and
                  Nora Kassner and
                  Elena Gribovskaya and
                  Sebastian Riedel and
                  Mor Geva},
  title        = {Do Large Language Models Perform Latent Multi-Hop Reasoning without
                  Exploiting Shortcuts?},
  year         = {2024},
  url          = {https://doi.org/10.48550/arXiv.2411.16679},
  archivePrefix= {arXiv},
  eprint       = {2411.16679}
}

@inproceedings{Chen24,
  author       = {Jiaao Chen and
                  Xiaoman Pan and
                  Dian Yu and
                  Kaiqiang Song and
                  Xiaoyang Wang and
                  Dong Yu and
                  Jianshu Chen},
  editor       = {Yaser Al{-}Onaizan and
                  Mohit Bansal and
                  Yun{-}Nung Chen},
  title        = {Skills-in-Context: Unlocking Compositionality in Large Language Models},
  booktitle    = {Findings of the Association for Computational Linguistics: {EMNLP}
                  2024, Miami, Florida, USA, November 12-16, 2024},
  pages        = {13838--13890},
  publisher    = {Association for Computational Linguistics},
  year         = {2024},
  url          = {https://aclanthology.org/2024.findings-emnlp.812}
}

@inproceedings{Li24chain,
  author       = {Chengshu Li and
                  Jacky Liang and
                  Andy Zeng and
                  Xinyun Chen and
                  Karol Hausman and
                  Dorsa Sadigh and
                  Sergey Levine and
                  Li Fei{-}Fei and
                  Fei Xia and
                  Brian Ichter},
  title        = {Chain of Code: Reasoning with a Language Model-Augmented Code Emulator},
  booktitle    = {Forty-first International Conference on Machine Learning, {ICML} 2024,
                  Vienna, Austria, July 21-27, 2024},
  publisher    = {OpenReview.net},
  year         = {2024},
  url          = {https://openreview.net/forum?id=vKtomqlSxm}
}

@misc{gandhi24,
  author       = {Kanishk Gandhi and
                  Denise Lee and
                  Gabriel Grand and
                  Muxin Liu and
                  Winson Cheng and
                  Archit Sharma and
                  Noah D. Goodman},
  title        = {Stream of Search (SoS): Learning to Search in Language},
  year         = {2024},
  url          = {https://doi.org/10.48550/arXiv.2404.03683},
  archivePrefix= {arXiv},
  eprint       = {2404.03683}
}

@inproceedings{Laskin23,
  author       = {Michael Laskin and
                  Luyu Wang and
                  Junhyuk Oh and
                  Emilio Parisotto and
                  Stephen Spencer and
                  Richie Steigerwald and
                  DJ Strouse and
                  Steven Stenberg Hansen and
                  Angelos Filos and
                  Ethan Brooks and
                  Maxime Gazeau and
                  Himanshu Sahni and
                  Satinder Singh and
                  Volodymyr Mnih},
  title        = {In-context Reinforcement Learning with Algorithm Distillation},
  booktitle    = {The Eleventh International Conference on Learning Representations,
                  {ICLR} 2023, Kigali, Rwanda, May 1-5, 2023},
  publisher    = {OpenReview.net},
  year         = {2023},
  url          = {https://openreview.net/forum?id=hy0a5MMPUv}
}

@misc{paglieri24,
  author       = {Davide Paglieri and
                  Bartlomiej Cupial and
                  Samuel Coward and
                  Ulyana Piterbarg and
                  Maciej Wolczyk and
                  Akbir Khan and
                  Eduardo Pignatelli and
                  Lukasz Kucinski and
                  Lerrel Pinto and
                  Rob Fergus and
                  Jakob Nicolaus Foerster and
                  Jack Parker{-}Holder and
                  Tim Rockt{\"{a}}schel},
  title        = {{BALROG:} Benchmarking Agentic {LLM} and {VLM} Reasoning On Games},
  year         = {2024},
  url          = {https://doi.org/10.48550/arXiv.2411.13543},
  archivePrefix= {arXiv},
  eprint       = {2411.13543}
}

@inproceedings{Allen23,
  author       = {Zeyuan Allen{-}Zhu and
                  Yuanzhi Li},
  title        = {Physics of Language Models: Part 3.2, Knowledge Manipulation},
  booktitle    = {The Thirteenth International Conference on Learning Representations,
                  {ICLR} 2025, Singapore, April 24-28, 2025},
  publisher    = {OpenReview.net},
  year         = {2025},
  url          = {https://openreview.net/forum?id=oDbiL9CLoS},
  bibsource    = {dblp computer science bibliography, https://dblp.org}
}

@inproceedings{Treutlein24,
  author       = {Johannes Treutlein and
                  Dami Choi and
                  Jan Betley and
                  Samuel Marks and
                  Cem Anil and
                  Roger B. Grosse and
                  Owain Evans},
  editor       = {Amir Globersons and
                  Lester Mackey and
                  Danielle Belgrave and
                  Angela Fan and
                  Ulrich Paquet and
                  Jakub M. Tomczak and
                  Cheng Zhang},
  title        = {Connecting the Dots: LLMs can Infer and Verbalize Latent Structure
                  from Disparate Training Data},
  booktitle    = {Advances in Neural Information Processing Systems 38: Annual Conference
                  on Neural Information Processing Systems 2024, NeurIPS 2024, Vancouver,
                  BC, Canada, December 10 - 15, 2024},
  year         = {2024},
  url          = {http://papers.nips.cc/paper\_files/paper/2024/hash/fe489a28a54583ee802b8e2955c024c2-Abstract-Conference.html}
}

@misc{shao2024deepseekmathpushinglimitsmathematical,
      title={DeepSeekMath: Pushing the Limits of Mathematical Reasoning in Open Language Models}, 
      author={Zhihong Shao and Peiyi Wang and Qihao Zhu and Runxin Xu and Junxiao Song and Xiao Bi and Haowei Zhang and Mingchuan Zhang and Y. K. Li and Y. Wu and Daya Guo},
      year={2024},
      eprint={2402.03300},
      archivePrefix={arXiv},
      primaryClass={cs.CL},
      url={https://arxiv.org/abs/2402.03300}, 
}

@article{llama3,
  author       = {Abhimanyu Dubey and
                  Abhinav Jauhri and
                  Abhinav Pandey and
                  Abhishek Kadian and
                  Ahmad Al{-}Dahle and
                  Aiesha Letman and
                  Akhil Mathur and
                  Alan Schelten and
                  Amy Yang and
                  Angela Fan and
                  Anirudh Goyal and
                  Anthony Hartshorn and
                  Aobo Yang and
                  Archi Mitra and
                  Archie Sravankumar and
                  Artem Korenev and
                  Arthur Hinsvark and
                  Arun Rao and
                  Aston Zhang and
                  Aur{\'{e}}lien Rodriguez and
                  Austen Gregerson and
                  Ava Spataru and
                  Baptiste Rozi{\`{e}}re and
                  Bethany Biron and
                  Binh Tang and
                  Bobbie Chern and
                  Charlotte Caucheteux and
                  Chaya Nayak and
                  Chloe Bi and
                  Chris Marra and
                  Chris McConnell and
                  Christian Keller and
                  Christophe Touret and
                  Chunyang Wu and
                  Corinne Wong and
                  Cristian Canton Ferrer and
                  Cyrus Nikolaidis and
                  Damien Allonsius and
                  Daniel Song and
                  Danielle Pintz and
                  Danny Livshits and
                  David Esiobu and
                  Dhruv Choudhary and
                  Dhruv Mahajan and
                  Diego Garcia{-}Olano and
                  Diego Perino and
                  Dieuwke Hupkes and
                  Egor Lakomkin and
                  Ehab AlBadawy and
                  Elina Lobanova and
                  Emily Dinan and
                  Eric Michael Smith and
                  Filip Radenovic and
                  Frank Zhang and
                  Gabriel Synnaeve and
                  Gabrielle Lee and
                  Georgia Lewis Anderson and
                  Graeme Nail and
                  Gr{\'{e}}goire Mialon and
                  Guan Pang and
                  Guillem Cucurell and
                  Hailey Nguyen and
                  Hannah Korevaar and
                  Hu Xu and
                  Hugo Touvron and
                  Iliyan Zarov and
                  Imanol Arrieta Ibarra and
                  Isabel M. Kloumann and
                  Ishan Misra and
                  Ivan Evtimov and
                  Jade Copet and
                  Jaewon Lee and
                  Jan Geffert and
                  Jana Vranes and
                  Jason Park and
                  Jay Mahadeokar and
                  Jeet Shah and
                  Jelmer van der Linde and
                  Jennifer Billock and
                  Jenny Hong and
                  Jenya Lee and
                  Jeremy Fu and
                  Jianfeng Chi and
                  Jianyu Huang and
                  Jiawen Liu and
                  Jie Wang and
                  Jiecao Yu and
                  Joanna Bitton and
                  Joe Spisak and
                  Jongsoo Park and
                  Joseph Rocca and
                  Joshua Johnstun and
                  Joshua Saxe and
                  Junteng Jia and
                  Kalyan Vasuden Alwala and
                  Kartikeya Upasani and
                  Kate Plawiak and
                  Ke Li and
                  Kenneth Heafield and
                  Kevin Stone and
                  et al.},
  title        = {The Llama 3 Herd of Models},
  journal      = {CoRR},
  volume       = {abs/2407.21783},
  year         = {2024},
  url          = {https://doi.org/10.48550/arXiv.2407.21783},
  doi          = {10.48550/ARXIV.2407.21783},
  eprinttype    = {arXiv},
  eprint       = {2407.21783}
}

@misc{zaremba2015learningexecute,
      title={Learning to Execute}, 
      author={Wojciech Zaremba and Ilya Sutskever},
      year={2015},
      eprint={1410.4615},
      archivePrefix={arXiv},
      primaryClass={cs.NE},
      url={https://arxiv.org/abs/1410.4615}, 
}

@misc{qwen2025qwen25technicalreport,
      title={Qwen2.5 Technical Report}, 
      author={Qwen and : and An Yang and Baosong Yang and Beichen Zhang and Binyuan Hui and Bo Zheng and Bowen Yu and Chengyuan Li and Dayiheng Liu and Fei Huang and Haoran Wei and Huan Lin and Jian Yang and Jianhong Tu and Jianwei Zhang and Jianxin Yang and Jiaxi Yang and Jingren Zhou and Junyang Lin and Kai Dang and Keming Lu and Keqin Bao and Kexin Yang and Le Yu and Mei Li and Mingfeng Xue and Pei Zhang and Qin Zhu and Rui Men and Runji Lin and Tianhao Li and Tianyi Tang and Tingyu Xia and Xingzhang Ren and Xuancheng Ren and Yang Fan and Yang Su and Yichang Zhang and Yu Wan and Yuqiong Liu and Zeyu Cui and Zhenru Zhang and Zihan Qiu},
      year={2025},
      eprint={2412.15115},
      archivePrefix={arXiv},
      primaryClass={cs.CL},
      url={https://arxiv.org/abs/2412.15115}, 
}

@misc{openai2024gpt4ocard,
      title={GPT-4o System Card}, 
      author={OpenAI and : and Aaron Hurst and Adam Lerer and Adam P. Goucher and Adam Perelman and Aditya Ramesh and Aidan Clark and AJ Ostrow and Akila Welihinda and Alan Hayes and Alec Radford and Aleksander Mądry and Alex Baker-Whitcomb and Alex Beutel and Alex Borzunov and Alex Carney and Alex Chow and Alex Kirillov and Alex Nichol and Alex Paino and Alex Renzin and Alex Tachard Passos and Alexander Kirillov and Alexi Christakis and Alexis Conneau and Ali Kamali and Allan Jabri and Allison Moyer and Allison Tam and Amadou Crookes and Amin Tootoochian and Amin Tootoonchian and Ananya Kumar and Andrea Vallone and Andrej Karpathy and Andrew Braunstein and Andrew Cann and Andrew Codispoti and Andrew Galu and Andrew Kondrich and Andrew Tulloch and Andrey Mishchenko and Angela Baek and Angela Jiang and Antoine Pelisse and Antonia Woodford and Anuj Gosalia and Arka Dhar and Ashley Pantuliano and Avi Nayak and Avital Oliver and Barret Zoph and Behrooz Ghorbani and Ben Leimberger and Ben Rossen and Ben Sokolowsky and Ben Wang and Benjamin Zweig and Beth Hoover and Blake Samic and Bob McGrew and Bobby Spero and Bogo Giertler and Bowen Cheng and Brad Lightcap and Brandon Walkin and Brendan Quinn and Brian Guarraci and Brian Hsu and Bright Kellogg and Brydon Eastman and Camillo Lugaresi and Carroll Wainwright and Cary Bassin and Cary Hudson and Casey Chu and Chad Nelson and Chak Li and Chan Jun Shern and Channing Conger and Charlotte Barette and Chelsea Voss and Chen Ding and Cheng Lu and Chong Zhang and Chris Beaumont and Chris Hallacy and Chris Koch and Christian Gibson and Christina Kim and Christine Choi and Christine McLeavey and Christopher Hesse and Claudia Fischer and Clemens Winter and Coley Czarnecki and Colin Jarvis and Colin Wei and Constantin Koumouzelis and Dane Sherburn and Daniel Kappler and Daniel Levin and Daniel Levy and David Carr and David Farhi and David Mely and David Robinson and David Sasaki and Denny Jin and Dev Valladares and Dimitris Tsipras and Doug Li and Duc Phong Nguyen and Duncan Findlay and Edede Oiwoh and Edmund Wong and Ehsan Asdar and Elizabeth Proehl and Elizabeth Yang and Eric Antonow and Eric Kramer and Eric Peterson and Eric Sigler and Eric Wallace and Eugene Brevdo and Evan Mays and Farzad Khorasani and Felipe Petroski Such and Filippo Raso and Francis Zhang and Fred von Lohmann and Freddie Sulit and Gabriel Goh and Gene Oden and Geoff Salmon and Giulio Starace and Greg Brockman and Hadi Salman and Haiming Bao and Haitang Hu and Hannah Wong and Haoyu Wang and Heather Schmidt and Heather Whitney and Heewoo Jun and Hendrik Kirchner and Henrique Ponde de Oliveira Pinto and Hongyu Ren and Huiwen Chang and Hyung Won Chung and Ian Kivlichan and Ian O'Connell and Ian O'Connell and Ian Osband and Ian Silber and Ian Sohl and Ibrahim Okuyucu and Ikai Lan and Ilya Kostrikov and Ilya Sutskever and Ingmar Kanitscheider and Ishaan Gulrajani and Jacob Coxon and Jacob Menick and Jakub Pachocki and James Aung and James Betker and James Crooks and James Lennon and Jamie Kiros and Jan Leike and Jane Park and Jason Kwon and Jason Phang and Jason Teplitz and Jason Wei and Jason Wolfe and Jay Chen and Jeff Harris and Jenia Varavva and Jessica Gan Lee and Jessica Shieh and Ji Lin and Jiahui Yu and Jiayi Weng and Jie Tang and Jieqi Yu and Joanne Jang and Joaquin Quinonero Candela and Joe Beutler and Joe Landers and Joel Parish and Johannes Heidecke and John Schulman and Jonathan Lachman and Jonathan McKay and Jonathan Uesato and Jonathan Ward and Jong Wook Kim and Joost Huizinga and Jordan Sitkin and Jos Kraaijeveld and Josh Gross and Josh Kaplan and Josh Snyder and Joshua Achiam and Joy Jiao and Joyce Lee and Juntang Zhuang and Justyn Harriman and Kai Fricke and Kai Hayashi and Karan Singhal and Katy Shi and Kavin Karthik and Kayla Wood and Kendra Rimbach and Kenny Hsu and Kenny Nguyen and Keren Gu-Lemberg and Kevin Button and Kevin Liu and Kiel Howe and Krithika Muthukumar and Kyle Luther and Lama Ahmad and Larry Kai and Lauren Itow and Lauren Workman and Leher Pathak and Leo Chen and Li Jing and Lia Guy and Liam Fedus and Liang Zhou and Lien Mamitsuka and Lilian Weng and Lindsay McCallum and Lindsey Held and Long Ouyang and Louis Feuvrier and Lu Zhang and Lukas Kondraciuk and Lukasz Kaiser and Luke Hewitt and Luke Metz and Lyric Doshi and Mada Aflak and Maddie Simens and Madelaine Boyd and Madeleine Thompson and Marat Dukhan and Mark Chen and Mark Gray and Mark Hudnall and Marvin Zhang and Marwan Aljubeh and Mateusz Litwin and Matthew Zeng and Max Johnson and Maya Shetty and Mayank Gupta and Meghan Shah and Mehmet Yatbaz and Meng Jia Yang and Mengchao Zhong and Mia Glaese and Mianna Chen and Michael Janner and Michael Lampe and Michael Petrov and Michael Wu and Michele Wang and Michelle Fradin and Michelle Pokrass and Miguel Castro and Miguel Oom Temudo de Castro and Mikhail Pavlov and Miles Brundage and Miles Wang and Minal Khan and Mira Murati and Mo Bavarian and Molly Lin and Murat Yesildal and Nacho Soto and Natalia Gimelshein and Natalie Cone and Natalie Staudacher and Natalie Summers and Natan LaFontaine and Neil Chowdhury and Nick Ryder and Nick Stathas and Nick Turley and Nik Tezak and Niko Felix and Nithanth Kudige and Nitish Keskar and Noah Deutsch and Noel Bundick and Nora Puckett and Ofir Nachum and Ola Okelola and Oleg Boiko and Oleg Murk and Oliver Jaffe and Olivia Watkins and Olivier Godement and Owen Campbell-Moore and Patrick Chao and Paul McMillan and Pavel Belov and Peng Su and Peter Bak and Peter Bakkum and Peter Deng and Peter Dolan and Peter Hoeschele and Peter Welinder and Phil Tillet and Philip Pronin and Philippe Tillet and Prafulla Dhariwal and Qiming Yuan and Rachel Dias and Rachel Lim and Rahul Arora and Rajan Troll and Randall Lin and Rapha Gontijo Lopes and Raul Puri and Reah Miyara and Reimar Leike and Renaud Gaubert and Reza Zamani and Ricky Wang and Rob Donnelly and Rob Honsby and Rocky Smith and Rohan Sahai and Rohit Ramchandani and Romain Huet and Rory Carmichael and Rowan Zellers and Roy Chen and Ruby Chen and Ruslan Nigmatullin and Ryan Cheu and Saachi Jain and Sam Altman and Sam Schoenholz and Sam Toizer and Samuel Miserendino and Sandhini Agarwal and Sara Culver and Scott Ethersmith and Scott Gray and Sean Grove and Sean Metzger and Shamez Hermani and Shantanu Jain and Shengjia Zhao and Sherwin Wu and Shino Jomoto and Shirong Wu and Shuaiqi and Xia and Sonia Phene and Spencer Papay and Srinivas Narayanan and Steve Coffey and Steve Lee and Stewart Hall and Suchir Balaji and Tal Broda and Tal Stramer and Tao Xu and Tarun Gogineni and Taya Christianson and Ted Sanders and Tejal Patwardhan and Thomas Cunninghman and Thomas Degry and Thomas Dimson and Thomas Raoux and Thomas Shadwell and Tianhao Zheng and Todd Underwood and Todor Markov and Toki Sherbakov and Tom Rubin and Tom Stasi and Tomer Kaftan and Tristan Heywood and Troy Peterson and Tyce Walters and Tyna Eloundou and Valerie Qi and Veit Moeller and Vinnie Monaco and Vishal Kuo and Vlad Fomenko and Wayne Chang and Weiyi Zheng and Wenda Zhou and Wesam Manassra and Will Sheu and Wojciech Zaremba and Yash Patil and Yilei Qian and Yongjik Kim and Youlong Cheng and Yu Zhang and Yuchen He and Yuchen Zhang and Yujia Jin and Yunxing Dai and Yury Malkov},
      year={2024},
      eprint={2410.21276},
      archivePrefix={arXiv},
      primaryClass={cs.CL},
      url={https://arxiv.org/abs/2410.21276}, 
}

@misc{lampinen2025generalization,
      title={On the generalization of language models from in-context learning and finetuning: a controlled study}, 
      author={Andrew K. Lampinen and
             Arslan Chaudhry and
             Stephanie C.Y. Chan and
             Cody Wild and
             Diane Wan and
             Alex Ku and
             Jörg Bornschein and
             Razvan Pascanu and
             Murray Shanahan and
             James L. McClelland},
      year={2025},
      eprint={2505.00661},
      archivePrefix={arXiv},
      primaryClass={cs.CL},
      url={https://arxiv.org/abs/2505.00661}, 
}

@article{Dienes1999Theory, 
    author={Zoltan Dienes and Josef Perner},
    title={A theory of implicit and explicit knowledge}, volume={22}, 
    number={5}, 
    journal={Behavioral and Brain Sciences}, 
    year={1999}, 
    pages={735–808}
}

@inproceedings{Zelikman2022Star,
  author       = {Eric Zelikman and
                  Yuhuai Wu and
                  Jesse Mu and
                  Noah D. Goodman},
  editor       = {Sanmi Koyejo and
                  S. Mohamed and
                  A. Agarwal and
                  Danielle Belgrave and
                  K. Cho and
                  A. Oh},
  title        = {STaR: Bootstrapping Reasoning With Reasoning},
  booktitle    = {Advances in Neural Information Processing Systems 35: Annual Conference
                  on Neural Information Processing Systems 2022, NeurIPS 2022, New Orleans,
                  LA, USA, November 28 - December 9, 2022},
  year         = {2022},
  url          = {http://papers.nips.cc/paper\_files/paper/2022/hash/639a9a172c044fbb64175b5fad42e9a5-Abstract-Conference.html}
}

@article{Qwen3,
  author       = {An Yang and
                  Anfeng Li and
                  Baosong Yang and
                  Beichen Zhang and
                  Binyuan Hui and
                  Bo Zheng and
                  Bowen Yu and
                  Chang Gao and
                  Chengen Huang and
                  Chenxu Lv and
                  Chujie Zheng and
                  Dayiheng Liu and
                  Fan Zhou and
                  Fei Huang and
                  Feng Hu and
                  Hao Ge and
                  Haoran Wei and
                  Huan Lin and
                  Jialong Tang and
                  Jian Yang and
                  Jianhong Tu and
                  Jianwei Zhang and
                  Jian Yang and
                  Jiaxi Yang and
                  Jingren Zhou and
                  Junyang Lin and
                  Kai Dang and
                  Keqin Bao and
                  Kexin Yang and
                  Le Yu and
                  Lianghao Deng and
                  Mei Li and
                  Mingfeng Xue and
                  Mingze Li and
                  Pei Zhang and
                  Peng Wang and
                  Qin Zhu and
                  Rui Men and
                  Ruize Gao and
                  Shixuan Liu and
                  Shuang Luo and
                  Tianhao Li and
                  Tianyi Tang and
                  Wenbiao Yin and
                  Xingzhang Ren and
                  Xinyu Wang and
                  Xinyu Zhang and
                  Xuancheng Ren and
                  Yang Fan and
                  Yang Su and
                  Yichang Zhang and
                  Yinger Zhang and
                  Yu Wan and
                  Yuqiong Liu and
                  Zekun Wang and
                  Zeyu Cui and
                  Zhenru Zhang and
                  Zhipeng Zhou and
                  Zihan Qiu},
  title        = {Qwen3 Technical Report},
  journal      = {CoRR},
  volume       = {abs/2505.09388},
  year         = {2025},
  url          = {https://doi.org/10.48550/arXiv.2505.09388},
  eprinttype    = {arXiv},
  eprint       = {2505.09388}
}

@inproceedings{Tian2019Learning,
  author       = {Yonglong Tian and
                  Andrew Luo and
                  Xingyuan Sun and
                  Kevin Ellis and
                  William T. Freeman and
                  Joshua B. Tenenbaum and
                  Jiajun Wu},
  title        = {Learning to Infer and Execute 3D Shape Programs},
  booktitle    = {7th International Conference on Learning Representations, {ICLR} 2019,
                  New Orleans, LA, USA, May 6-9, 2019},
  year         = {2019},
  url          = {https://openreview.net/forum?id=rylNH20qFQ}
}

@inproceedings{Yan2020Neural,
  author       = {Yujun Yan and
                  Kevin Swersky and
                  Danai Koutra and
                  Parthasarathy Ranganathan and
                  Milad Hashemi},
  title        = {Neural Execution Engines: Learning to Execute Subroutines},
  booktitle    = {Advances in Neural Information Processing Systems 33: Annual Conference
                  on Neural Information Processing Systems 2020, NeurIPS 2020, December
                  6-12, 2020, virtual},
  year         = {2020},
  url          = {https://proceedings.neurips.cc/paper/2020/hash/c8b9abffb45bf79a630fb613dcd23449-Abstract.html}
}

@inproceedings{Rahaman2021Dynamic,
  author       = {Nasim Rahaman and
                  Muhammad Waleed Gondal and
                  Shruti Joshi and
                  Peter V. Gehler and
                  Yoshua Bengio and
                  Francesco Locatello and
                  Bernhard Sch{\"{o}}lkopf},
  title        = {Dynamic Inference with Neural Interpreters},
  booktitle    = {Advances in Neural Information Processing Systems 34: Annual Conference
                  on Neural Information Processing Systems 2021, NeurIPS 2021, December
                  6-14, 2021, virtual},
  pages        = {10985--10998},
  year         = {2021},
  url          = {https://proceedings.neurips.cc/paper/2021/hash/5b4e9aa703d0bfa11041debaa2d1b633-Abstract.html}
}

@article{Lampinen2025Latent,
  author       = {Andrew Kyle Lampinen and
                  Martin Engelcke and
                  Yuxuan Li and
                  Arslan Chaudhry and
                  James L. McClelland},
  title        = {Latent learning: episodic memory complements parametric learning by
                  enabling flexible reuse of experiences},
  journal      = {CoRR},
  volume       = {abs/2509.16189},
  year         = {2025},
  url          = {https://doi.org/10.48550/arXiv.2509.16189},
  eprinttype    = {arXiv},
  eprint       = {2509.16189}
}

@inproceedings{Ou2025How,
  author       = {Yixin Ou and
                  Yunzhi Yao and
                  Ningyu Zhang and
                  Hui Jin and
                  Jiacheng Sun and
                  Shumin Deng and
                  Zhenguo Li and
                  Huajun Chen},
  editor       = {Wanxiang Che and
                  Joyce Nabende and
                  Ekaterina Shutova and
                  Mohammad Taher Pilehvar},
  title        = {How Do LLMs Acquire New Knowledge? {A} Knowledge Circuits Perspective
                  on Continual Pre-Training},
  booktitle    = {Findings of the Association for Computational Linguistics, {ACL} 2025,
                  Vienna, Austria, July 27 - August 1, 2025},
  pages        = {19889--19913},
  publisher    = {Association for Computational Linguistics},
  year         = {2025},
  url          = {https://aclanthology.org/2025.findings-acl.1021/}
}

@article{Cagatay2025Investigating,
  author       = {{\c{C}}agatay Yildiz and
                  Nishaanth Kanna Ravichandran and
                  Nitin Sharma and
                  Matthias Bethge and
                  Beyza Ermis},
  title        = {Investigating Continual Pretraining in Large Language Models: Insights
                  and Implications},
  journal      = {Trans. Mach. Learn. Res.},
  volume       = {2025},
  year         = {2025},
  url          = {https://openreview.net/forum?id=aKjJoEVKgO}
}

@article{Parmar2024Reuse,
  author       = {Jupinder Parmar and
                  Sanjeev Satheesh and
                  Mostofa Patwary and
                  Mohammad Shoeybi and
                  Bryan Catanzaro},
  title        = {Reuse, Don't Retrain: {A} Recipe for Continued Pretraining of
                  Language Models},
  journal      = {CoRR},
  volume       = {abs/2407.07263},
  year         = {2024},
  url          = {https://doi.org/10.48550/arXiv.2407.07263},
  doi          = {10.48550/ARXIV.2407.07263},
  eprinttype    = {arXiv},
  eprint       = {2407.07263}
}

@inproceedings{Chen2025Towards,
  author       = {Jie Chen and
                  Zhipeng Chen and
                  Jiapeng Wang and
                  Kun Zhou and
                  Yutao Zhu and
                  Jinhao Jiang and
                  Yingqian Min and
                  Xin Zhao and
                  Zhicheng Dou and
                  Jiaxin Mao and
                  Yankai Lin and
                  Ruihua Song and
                  Jun Xu and
                  Xu Chen and
                  Rui Yan and
                  Zhewei Wei and
                  Di Hu and
                  Wenbing Huang and
                  Ji{-}Rong Wen},
  editor       = {Wanxiang Che and
                  Joyce Nabende and
                  Ekaterina Shutova and
                  Mohammad Taher Pilehvar},
  title        = {Towards Effective and Efficient Continual Pre-training of Large Language
                  Models},
  booktitle    = {Proceedings of the 63rd Annual Meeting of the Association for Computational
                  Linguistics (Volume 1: Long Papers), {ACL} 2025, Vienna, Austria,
                  July 27 - August 1, 2025},
  pages        = {5779--5795},
  publisher    = {Association for Computational Linguistics},
  year         = {2025},
  url          = {https://aclanthology.org/2025.acl-long.289/}
}

@book{Jones1993Partial,
  author       = {Neil D. Jones and
                  Carsten K. Gomard and
                  Peter Sestoft},
  title        = {Partial evaluation and automatic program generation},
  series       = {Prentice Hall international series in computer science},
  publisher    = {Prentice Hall},
  year         = {1993},
  isbn         = {978-0-13-020249-9}
}
